%% file: emnlp2023.tex
\pdfoutput=1

\documentclass[11pt]{article}

\usepackage{EMNLP2023}

\usepackage{times}
\usepackage{latexsym}

\usepackage[T1]{fontenc}

\usepackage[utf8]{inputenc}

\usepackage{microtype}
\usepackage{amsmath} 
\usepackage{inconsolata}
\usepackage{tcolorbox}
\usepackage{adjustbox}

\usepackage{placeins}

\usepackage{xspace}
\newcommand{\thickhline}{\noalign{\hrule height 1pt}}
\newcommand{\ours}{SumAutoEval\xspace}
\newcommand{\Organization}{Alignment\xspace}
\title{Evaluate Summarization in Fine-Granularity: Auto Evaluation with LLM}

\author{%
Dong Yuan* ~~~~ Eti Rastogi* \\
\textit{DeepScribe Inc.}\\
\textit{doffery20@gmail.com, eti\_rastogi@yahoo.com}\\
\\
\textbf{Fen Zhao ~~ Sagar Goyal ~~ Gautam Naik ~~ Sree Prasanna Rajagopal} \\ 
\textit{DeepScribe Inc.}\\
\textit{\{sagar, fen, gautam, sree\}@deepscribe.tech}\\
}

\begin{document}
\maketitle
\footnotetext[1]{*Work done during work in DeepScribe.}
\begin{abstract}
Due to the exponential growth of information and the need for efficient information consumption the task of summarization has gained paramount importance. Evaluating summarization accurately and objectively presents significant challenges, particularly when dealing with long and unstructured texts rich in content. Existing methods, such as ROUGE \citep{lin-2004-rouge} and embedding similarities, often yield scores that have low correlation with human judgements and are also not intuitively understandable, making it difficult to gauge the true quality of the summaries. LLMs can mimic human in giving subjective reviews but subjective scores are hard to interpret and justify. They can be easily manipulated by altering the models and the tones of the prompts. In this paper, we introduce a novel evaluation methodology and tooling designed to address these challenges, providing a more comprehensive, accurate and interpretable assessment of summarization outputs. Our method (\ours) proposes and evaluates metrics at varying granularity levels, giving objective scores on 4 key dimensions such as completeness, correctness, \Organization and readability. We empirically demonstrate, that \ours enhances the understanding of output quality with better human correlation.

\end{abstract}

\section{Introduction}
The LLMs, e.g. GPT-4, Claude, exceeded the summarization capabilities previously known. These models can generate high-quality summaries which are indistinguishable from human-written texts but also often suffer from missing information, hallucinations, misinterpretation of facts and various other forms of inaccuracies which are hard tYesah o identify and measure by both machine and human.



The unstructured nature of the text outputs for summarization task makes evaluation a specifically challenging task. Many tools like ROUGE score \citep{lin-2004-rouge}, embedding based cosine similarity have been used in the past but they tend to be non-intuitive score numbers, highly variant and most importantly correlates badly with human judgements. In LLM generated summarization world, these metrics are ineffcient in detecting missing details and hallucinations errors.
The use of these metrics in critical domain-specific use cases such as medical note summarization can be even more concerning as the model can miss a lot of relevant information and still have good metrics. 


Recently, more LLM-based evaluation techniques \citep{liu-etal-2023-g} have emerged, aiming to replicate human ratings on summary quality and have even achieved strong correlations with human subjective ratings \citep{liu-etal-2023-g}. Building on the framework defined by \citep{kryscinski2019neural}, which includes Coherence, Consistency, Fluency, and Relevance as the four subjective dimensions for summarization evaluation, many subsequent works \citep{fabbri2021summeval, liu-etal-2023-g} have adopted these criteria. These works typically involve subjective ratings from 1 to 5 to assess summarization. However, these metrics are often seen as not sufficiently objective and concrete, making it challenging for models to excel and provide mathematically explainable results.


In this paper, we propose a new set of metrics that are both machine-readable and reflective of the four key subjective aspects of summarization in a measurable manner. Additionally, we introduce an innovative LLM-based framework designed to evaluate these metrics effectively. Our novel framework is designed to objectively evaluate the quality of summarized text based on four key pillars: \textbf{Completeness, Correctness, Organization, and Readability}. Much like the process of scoring an answer sheet according to a precise answer key, our method ensures that each aspect of the text is measured in an objective and quantifiable manner. This approach not only provides a clear understanding of the summarized content's quality but also eliminates bias, ensuring that the scores are stable and robust regardless of changes in prompts or models. By making our evaluation method both consistent and dependable, we achieve accuracy akin to an impartial human assessor, thereby maintaining score reliability even if the evaluator changes.


\input{section2.0-metrics}

\input{Section2.2-method}

\input{section5.0-evaluation_setup}

\input{section5.2-medical}

\input{section5.1-summeval}

\input{section6-related-work}

\section{Conclusion \& Future Work}
This paper proposed a fine-granularity evaluation method on summarization output and proves its advantage over existing methods in capturing details and better alignment with humans in several dimensions. Later it's worth exploring how to combine the fine-granular way and overall evaluation into a single solution to generate a more comprehensive evaluation of a summarization note.





\bibliographystyle{acl_natbib}
\bibliography{anthology,custom}

\appendix
\onecolumn

\section{Example Appendix}
\label{sec:appendix}
\input{section7-appendix}

\end{document}

%% file: section2.0-metrics.tex
\section{Metrics} \label{metrics}
To thoroughly evaluate a summary, we define four key aspects of quality. These metrics encompass objective criteria that are more machine-understandable as opposed to subjective measures.

We use the term \textit{source} to refer to the text that needs to be summarized, \textit{target note } to refer to the summary being evaluated, and  \textit{ground truth note}  to denote the ground-truth summary of the source, which is accurate, detailed, well-organized, and curated by expert humans.





\noindent {\bf Completeness}: This aspect measures the extent to which relevant information from the ground truth note is covered in the target note. It penalizes any relevant information from the ground truth note that is missing in the target note.

\noindent {\bf Correctness}: This metric evaluates the accuracy of information presented in the target note. It identifies and penalizes hallucinations, misinterpretations, incorrect facts, and other inaccuracies in the target note.

\noindent {\bf \Organization}: This criterion assesses how well the information in the target note aligns with that in the ground truth note. It penalizes irrelevant or extraneous information included in the target note. In medical note summarization, this metric is crucial because the mere presence of information is insufficient; it must also be placed in the correct section and under the right category. For example, referencing imaging scans like an X-ray is appropriate and aligned only if it is recorded under the Subjective-Imaging section and not elsewhere in the note.

\noindent {\bf Readability}: This metric grades the target note on professional writing quality. It is penalized by awkward sentence flow, grammatical errors, inappropriate language, and spelling mistakes.








As defined in SummEval~\citep{fabbri2021summeval}, summarization is broadly evaluated on four key metrics - Consistency, Relevance, Coherence and Fluency. However, in our observation these are not enough and needs to be broken down into more granular and measurable metrics. For instance, Relevance handedly covers two key parameters - missing information which we measure in Completeness and presence of extra or irrelevant information which is represented by \Organization in our metrics. Similarly, Coherence can be mapped to Readability-Awkward Flow, Fluency to Readability-Grmmar and Spelling Errors and Consistencuy to Correctness.

%% file: Section2.2-method.tex
\section{Method}
This section introduces the technical details of our evaluation method.
\subsection{Entity Extraction} \label{entity_extraction}
To extract key entities from a summary accurately, we designed a three-step solution.
We define an \textit{Entity} as a unit of information, typically a short phrase that is both concise and complete in its meaning. An Entity contains exactly one key concept that is essential to the summary and without which the summary would bre considered incomplete.

1. {\bf Extraction.} An initial step to break down the note into short phrases, emphasizing each phrase should have an intact meaning, and also only have one key point. For example, "A and B are both normal when C happens" will be two "A is normal when C happens" and "B is normal when C happens".

2. {\bf Self-Verification.} This step enables the model to verify its generated output and merge contextually similar entities. For instance, an isolated date holds little significance unless it is connected to another action or context-related entity.

3. {\bf Reference Sourcing.} This prompt assists in identifying the original phrases or sentences to support the entities. It represents the final phase of refinement, where unsupported entities are removed. Additionally, the phrases or sentences, \textit{Entity Reference}, which pertains to the word phrases from the original note, is used to calculate correlation with human inputs on entities.


\subsection{Metrics Calculation}
With target note and the extracted target note entities, ground truth note, and ground truth note entities, we collect entity level results using a unified prompt template for Completeness, Correctness, and \Organization.

\begin{quote}
\begin{verbatim}
System Message: [Identity] [Goal]
User Message:
[Task Description]
[Output Options]
[Guideline]
[Examples]
[Reference Materials]
[Question]
[Chain of Thoughts Request]
[Output Formatting]
\end{verbatim}
\end{quote}

Based on different evaluation tasks, \textit{Identity} and \textit{Goal} are set differently. For example, for medical note summarization task, \textit{Identity} is "You are a helpful assistant good reviewing medical note".
\textit{Task Description} gives description about the task, e.g. "verify the correctness of an entity by referencing the source transcript". \textit{Output Options} describes the possible output answers. \textit{Guideline} and \textit{Example} gives concrete interpretation of how to solve the task. \textit{Reference Materials} are either ground truth note, source or other material that model can use as a reference for answering the question.

Inspired by self-consistency prompt~\citep{wang2022self}, to further improve the accuracy of the metrics, for classification tasks, we design \textit{Consistency Prompts}, which has multiple different prompts to evaluate from different angles and then aggregate the answers to get more accurate results. The reason why we do not use self-consistency prompt to improve accuracy here, is from our practical experience, sending the same prompt to LLM, a lot of times they return the exact the same output, which does not improve the accuracy comparing to our \textit{Consistency Prompts}.

\subsubsection{Completeness} \label{method:missing}
To measure the missing ratio of the ground truth rubric entity to reflect note's completeness, we designed multiple (currently three) prompts to measure the "missing" property of an entity from the ground truth note.

\noindent Prompt 1. Checks whether a given entity “is present” in the target note or not.

\noindent Prompt 2. Analyzes the “concept coverage” of information from the entity, aka the information within entity is covered in the target note or not.

\noindent Prompt 3. Finds relevant materials from the target note to support the entity, can we find enough materials/information to support the information in the entity or not.

These prompts aim the same goal, but guide the model to analyze from different perspective to find out the answer.

However, this approach may incorrectly label information that is inaccurate or contradicts the ground truth as still being complete. To mitigate this issue, we include "partial" and "contradict" as options in the \textit{Output Option}. An example of output options definition:
\begin{quote}
Yes: The concept content is covered in the target note. \\
No: None of the key points in concept is covered in the target note. \\
Contradict: the target note mentioned relevant information related to the concept but contradicts or refutes it. \\
Partially: While some elements from the concept are included in the target note, there are also key pieces of information from the concept that have not been integrated.

\end{quote}
Each output option is accompanied by a detailed explanation that includes the model's step-by-step reasoning and justification for its answers, specifically in relation to each of the given reference entities.
\begin{quote}

\end{quote}
For all prompts, we aggregated the answers using a majority voting system for each entity, with specific exception rules. For instance, 'partial + partial + yes' is aggregated as 'yes.' If there is no agreement, the answer defaults to 'no.' We award 0.5 points to 'partial' responses and 1 point each for 'yes' responses to calculate the entity score

After all entities are evaluated, we calculate missing entity ratio on ground truth note to get completeness metrics as follows -

\begin{equation}
{\small
\text{Completeness Score} =  \frac{ \text{Entity Score in the target note}}{\text{\#Entities in the ground truth note}}
}
\end{equation}
More advanced technology, like using graphic modeling tech to model each prompt's quality and potential answer could be used to further improve this scoring metric.


\subsubsection{Correctness} 
Similar to completeness measurement, multiple prompts are designed to evaluate the target note's correctness from different perspectives. Both the ground truth note and source materials are provided within the prompt. The source materials serve to verify information that may not be covered in the ground truth but is still correct (e.g., additional details that aren't relevant enough to be included in the ground truth note). While the prompt design aligns with that of completeness measurement, the output options differ. They include "Yes," "No," "Partially," and "Unknown." The "Unknown" option is specifically designed to allow the model to more accurately capture cases where it cannot find any information to support or refute a given entity.
Majority voting is also applied here to get the final entity level answer. Based on these entity-level answers, the incorrectness ratio is calculated to derive the correctness score

\begin{equation}
{\small
\text{Correctness Score} =  \frac{ \text{Entity Score in the target note}}{\text{\#Entities in the target note}}
}
\end{equation}

\subsubsection{\Organization}
To evaluate whether an entity is relevant by determining if it’s correctly positioned in the appropriate section, we employ two different prompt designs. The first method involves checking by definition, utilizing the header, section name, or other predefined criteria to understand what should be included in that section. The second method involves verifying whether the entity appears in the corresponding section of the ground truth note. Both methods aim to identify irrelevant entities. Ideally, one of these prompts should work perfectly on its own, but in the real world, both tend to make mistakes and they compliment each other. 
Alignment score is generated by the ratio of correctly placed entities.

\subsubsection{Readability}
Readability concerns multiple aspects of the writing style of the target note. It includes awkward flow, inappropriate language, grammar, and spelling issue. 
For awkward flow, prompt is designed to analyze the relationship among sentences to identify the broken logic. For inappropriate language, grammar, and spelling issue, prompt is designed to directly identify these issues, as it's straightforward for a language model. 
After answer aggregation, we add up all errors to calculate metric scores for readability.


%% file: section5.0-evaluation_setup.tex
\section{Evaluation}

\subsection{Baselines}

\textbf{Rouge.} Rouge score measures the n-gram or longest common sequence similarity between two sequences on word level.

\noindent\textbf{BARTScore} \citep{yuan2021bartscore} It measures the conditional probability of one text generating another to evaluate the text similarity based on the BART model.

\noindent\textbf{G-Eval} \citep{liu-etal-2023-g} It prompts LLMs, e.g. GPT3.5, GPT-4, to evaluate the quality of summary to input source to give a 1-5 rating on Cosistency, Relevance, Coherence and Fluency.



%% file: section5.2-medical.tex
\begin{table*}[t]
\centering
\small
\begin{tabular}{c c c c c c c c c}
\thickhline
Model & \multicolumn{1}{c}{\textbf{Completeness}} & \multicolumn{1}{c}{\textbf{Correctness}} & \multicolumn{1}{c}{\textbf{Organization}} & \multicolumn{1}{c}{\textbf{Readability}} \\
\hline
Rouge-1 & 0.28 & 0.25 & - & - \\
Rouge-L & 0.24 & 0.20 & - & - \\
BARTScore & - & - & - & - \\
\hline
\textbf{\ours-Sonnet} & 0.59 & 0.41 & 0.52 & 0.47 \\
\textbf{\ours-GPT4o} & 0.81 & 0.54 & 0.60 & 0.47 \\
\thickhline
\end{tabular}
\caption {\textbf{Medical Entity Summarization Cohen's Kappa Agreement Comparison.}}
\vspace{-5pt}
\label{table:medical_entity_correlation}
\end{table*}


\begin{table*}[t]
\centering
\small
\begin{tabular}{c c c c c c c c c}
\thickhline
Model &  \multicolumn{2}{c}{\textbf{Organization}} & \multicolumn{2}{c}{\textbf{Readability}} \\
 & $\rho$ & $\tau$ & $\rho$ & $\tau$ \\
\hline
Rouge-1 & 0.55 & 0.58 & 0.04 & 0.04 \\
Rouge-L & 0.06 & 0.05 & \~0 & \~0 \\
BARTScore & 0.05 & 0.05 & 0.10 & 0.11 \\
G-Eval* & 0.92 & 0.90 & 0.27 & 0.26 \\
\hline
\textbf{\ours} & 0.98 & 0.95 & 0.05 & 0.06 \\
\thickhline
\end{tabular}
\caption {Medical Note Summarization Model \& Human correlation on Spearman ($\rho$) and Kendall-Tau ($\tau$)}
\vspace{-5pt}
\label{table:medical_correlation}
\end{table*}


\subsection{Medical Note Summarization}

\noindent\textbf{Medical Note Summarization Dataset}. We curated a medical note summarization dataset. The source is de-identified real doctor and patient conversation transcript data, summary is a note including Chief Compliant, Symptoms, Medications, Medical History, Family History, Surgical History Social History, Labs / Tests / Imaging etc. This dataset has 30 notes in total, the ground truth note is generated by human experts (with human errors). We use GPT-4-32k to generate note for these conversations. The generated note (target note) is reviewed by human against the ground truth note with both entity-level labeling and final review scores, note that this score is still subjective but it's based on the results of entity labeling.

First, we analyze the correlation between solutions with human on entity labeling results in Table~\ref{table:medical_entity_correlation}. Note that Rouge and BARTScore cannot measure Organization and Readability, so the comparison is done among SumAutoEval variations and G-Eval in Table~\ref{table:medical_entity_correlation} and \ref{table:medical_correlation}. 

We use Cohen's Kappa score (or agreement) to analyze the correlation of human and solution as for each entity it's similar to a classification task. Note that, we don't use precision and recall as human labels actually also have quite some errors. As shown in Table \ref{table:medical_entity_correlation}, \ours aligns with human better than Rouge and BARTScore solutions, this is because \ours uses LLM as backend, which has stronger capability in verifying entity's existence or correctness. Rouge and BARTScore does not cover the organization and readability as these two requires to do entity understanding instead of only entity comparison across ground truth and target note. We also observe that GPT-4o model aligns with human much better than Sonnet in completeness, and correctness organization as these metrics require reasoning capability. But for readability, it's more sensation of word and narratives, Sonnet model is as powerful as GPT-4o. Worth to note that, in Readability, the correlation between \ours and human is lower than G-Eval. That's probably because awkward flow is is the major error of readability. And capturing this at whole note level instead of sentences level has between alignment with human.



%% file: section5.1-summeval.tex
\subsection{SummEval}


\begin{table*}[t]
\centering
\small
\begin{tabular}{c c c c c c c c c}
\thickhline
Model & \multicolumn{2}{c}{\textbf{Consistency}} & \multicolumn{2}{c}{\textbf{Relevance}} & \multicolumn{2}{c}{\textbf{Coherence}} & \multicolumn{2}{c}{\textbf{Fluency}} \\
 & $\rho$ & $\tau$ & $\rho$ & $\tau$ & $\rho$ & $\tau$ & $\rho$ & $\tau$ \\
\hline
Rouge-L & - & - & - & - & - & - & - & - \\
InstructScore & - & - & - & - & - & - & - & - \\
G-Eval & 0.61 & 0.57 & 0.67 & 0.58 & 0.56 & 0.48  & 0.59 & 0.54\\
\indent\indent{-w/o outliers} & 0.62 & 0.58 & 0.67 & \textbf{0.59} & 0.60 & 0.52 & \textbf{0.63} & \textbf{0.57} \\
\hline
\textbf{Ours} & 0.65 & 0.60 & 0.54 & 0.42 & 0.6 & 0.49 & 0.51 & 0.45 \\
\indent\indent{-w/o outliers} &  \textbf{0.70} &  \textbf{0.65} & \textbf{0.70} & 0.58 & \textbf{0.75} & \textbf{0.62} & 0.54 & 0.48 \\
\thickhline
\end{tabular}
\caption {\textbf{Model \& Human correlation on Spearman ($\rho$) and Kendall-Tau ($\tau$)} on \space SummEval$^{\dagger}$ \ Dataset}
\vspace{-5pt}
\label{table:summ_eval}
\end{table*}

SummEval dataset \citep{fabbri2021summeval} is designed specifically for evaluating the performance of automatic text summarization systems. It contains human annotations and comprehensive benchmarks that help researchers and developers assess the quality of summaries produced by various algorithms.

To effectively evaluate the quality of machine-generated summaries against human-compiled summaries, we employed a structured approach involving rubric creation, expert evaluation, and metric-based analysis.


\noindent\textbf{Expert Score Consistency and Averaging.} We observed a notable variance in expert scores across different machine summaries. 
\textcolor{black}{To address the inconsistency in the original scores, we chose a machine summary (from the pool of summaries) where expert evaluations were more consistent. We then averaged these consistent expert scores to create a reliable baseline. This average score was used as a benchmark for future comparisons.} We refer to this new subset dataset as \textbf{SummEval}$^{\dagger}$ \textbf{Dataset}.

\noindent\textbf{Rubric Creation and Entity Filtering.} We created a rubric by aggregating entities from the given human summaries. To ensure the relevance and significance of the entities included in the rubric, we filtered out any entities that were not present in at least five human summaries. This step was critical in focusing our evaluation on consistently important information, as determined by human experts. We used gpt-4 as the base model to run all our prompts.


\noindent\textbf{Metrics.} In Table~\ref{table:summ_eval}, we report the correlation between experts and our autoEval ratings on several metrics: consistency, relevance, fluency and coherence. We also report the scores after excluding the outliers where we observed that the human score was erroneous as described in the Data Error Analysis section below. 

\textbf{\textit{Consistency}} To measure consistency, we focused on the "incorrect" metric, which closely aligns with the definition of consistency. The score was calculated using the formula:
\begin{equation}
{\small
\text{Consistency Score} =  \frac{ \text{Total Entities} - \text{Incorrect Entities}}{\text{Total Entities}}
}
\end{equation}
This score was then normalized to a scale ranging from 1 to 5.

\textbf{\textit{Relevance}} For relevance, we combined the "missing" and "irrelevant" metrics as the human experts were required to assess both the presence and relevance of entities. The percentage of found entities(partial or complete) in the rubric and the percentage of irrelevant entities in the test note were computed and averaged, providing a comprehensive measure of relevance. 

\textbf{\textit{Fluency and Coherence}} To evaluate fluency and coherence, we utilized our Writing Issue metric. Fluency was assessed by identifying grammar and spelling errors, while coherence was determined through the presence of awkward flow in sentences. The score was calculated by determining the number of sentences with issues and dividing it by the total number of sentences, thereby capturing the percentage of the note with issues:

\begin{equation}
{\small
\text{Writing Issue Score} = \frac{\text{Sentences with Issues}}{\text{Total Sentences}}
}
\end{equation}
This methodology ensured a robust and structured evaluation framework, allowing us to quantify various aspects of summary quality and draw meaningful comparisons between human and machine-generated summaries.

\subsubsection{Data Error Analysis}
\textbf{\textit{Relevance}}
\textcolor{black}{During the rubric creation process, we discovered two articles that lacked any common entities appearing in at least five human summaries, prompting us to exclude them from our calculations.} Additionally, we observed that in at least 10 instances the expert scores were not accurate.
Since the experts were given a single summary chosen at random as a reference, we believe it might have contained unique entities not aligning with our methodology, which relies on common entities to identify the main theme of the article. These 10 examples were substantial enough to reduce our metrics by \textcolor{black}{23\%}. See Appendix \ref{apdx_rel} for an example. \\ 
\textbf{\textit{Coherence}} We observed a pattern similar to what we found in relevance assessment: a few data samples had ratings that did not match the autoEval results. After further analysis and human evaluation, we realized that the experts' ratings often did not accurately reflect the coherence of the summaries. To substantiate our findings, we included a reference example in the appendix \ref{apdx_coh}. Notably, a small number of data points (around 10) were sufficient to reduce the correlation by \textcolor{black}{20\%}.\\ 
\textbf{\textit{Fluency}} Upon human analysis, we observed that most ratings disregarded capitalization errors and missing punctuation. Our methodology, developed to mimic a professional writer's standards, strongly penalizes errors in capitalization, lengthy sentences, and other similar issues. However, to maintain consistency with the definition of fluency as outlined in \cite{fabbri2021summeval}, we only included errors where there were serious grammar issues, such as missing subjects or extensive misuse/lack of punctuation, which made the summary difficult to comprehend. See Appendix \ref{apdx_flu} for reference.\\
\noindent \textbf{\textit{Consistency}} Similar to the analysis above, we also observed that some machine summaries, which contained inaccurate information or misinterpreted facts from the article, were mistakenly rated as 5. An example is provided in the Appendix \ref{apdx_consis} to illustrate this.


%% file: section6-related-work.tex
\section{Related Work}
\noindent\textbf{Rouge based Evaluation.} ROUGE \citep{lin-2004-rouge}, is a very popular evaluation metric and can be defined such that, ROUGE-N captures the similarity based off of exact overlap of N-grams. ROUGE-L is also often used, where L stands for the longest common sequence. 

\noindent\textbf{Embedding based Evaluation.} Methods like BERTScore \citep{zhang2020bertscore} and MoverScore \citep{zhao2019moverscore} asses the embedding based similarity of the produced text with the ground truth at different token granularity levels using BERT based embeddings. Sub-Sentence Encoder \citep{chen2023subsentence} also introduces a method to extract and embed atomic propositions from a paragraph. These embeddings can then be used to infer the extent of information capture. \citep{zhong2022towards} trains specialized classifiers to evaluate each dimension.


\noindent\textbf{LLM based Evaluation.}
There has been a lot of work~\citep{gao2024llm} in using LLM's for evaluation. \citep{gao2023humanlike} used ChatGPT to annotate the outputs using Human-like methods such as Likert Scoring, Pyramid and outperformed commonly used model based metrics. \citep{chan2023chateval} goes one step beyond to use muti-agent to discuss the evaluation. Research around training models specifically for evaluation has also shown promise - \citep{wang2023large} train a model to evaluate model outputs, whereas \citep{selfee2023} and \citep{wang2023chatgpt} train a critique LLM. More recent work {Prometheus} builds a 13B model which has high correlation with GPT4 on evaluation scores along multiple preference dimensions. Using GPT4 itself as an evaluator using prompting as also become popular as in Alpaca \citep{alpaca} and Vicuna \citep{zheng2023judging}. Recently G-Eval \citep{liu-etal-2023-g} uses CoT prompting on GPT4 to subjectively evaluate language generation tasks and measure the correlation with expert human judges.


\noindent\textbf{Long Form Text Evaluation}. Besides general generation output evaluation. Long-form output results, like long detail summarization, are particularly hard to evaluate, which needs more fine-granularity evaluation  \citep{xu2023critical}.
Currently most of them  \citep{min2023factscore, wei2024long} focus on how to verify the factual part of the output and using external knowledge source.



%% file: section7-appendix.tex
\subsection{Outlier Example for Relevance} 
\begin{tcolorbox}[colback=gray!30, colframe=black]
\label{apdx_rel}
\textbf{Article Id:} \\
c50d33e9749e7bb484d9b69c4f5fca35
a3a50cb5 \\ 
\textbf{Machine Summary:} jacob phillips , 23 , was found dead at the end of the disused northern 
promenade . he ran from the driver , who gave chase at the seaside town of penarth .
an inquest heard how the trio got out of the taxi to ` use an atm ' . \\
\textbf{Expert Scores:} 4, 4, 4 (avg - 4) \\
\textbf{AutoEvalScore:} 2 \\
\textbf{Reasoning:} As per the rubric, the main theme of the article is:
    1. Jacob Phillips, a 23-year-old accountant, fell to his death from a 70-foot cliff.
    2. The incident occurred in December after a night out with friends.
    3. He was running away from a taxi driver to avoid paying the fare.
    4. Phillips and his friends did not have enough money to pay for the taxi fare.
    
    However, the given machine summary completely misses the point that the young man died 
    because he was running away from a taxi driver to avoid paying the fare. The summary 
    doesn't clearly explain why Jacob was running from the driver and how he died.
\end{tcolorbox}

\subsection{Outlier Example for Coherence}
\begin{tcolorbox}[colback=gray!30, colframe=black]
\label{apdx_coh}
\textbf{Article Id:} \\
4761dc6d8bdf56b9ada97104113dd1bcf4aed3f1\\
\textbf{Machine Summary:} north pacific gray whale swam nearly 14,000 miles -lrb- 22,500 kilometers -rrb- . varvara , which is russian for `` barbara , '' left her primary feeding ground off russia 's sakhalin island . varvara 's journey surpassed a record listed on the guinness worlds records website . \\
\textbf{Expert Scores:} 4,4,4 (avg -4)\\
\textbf{AutoEvalScore:}  1\\
\textbf{Reasoning:}  \\
The summary lacks a coherent flow of information. It jumps from the whale's distance swam to the meaning of its name and then to the record it broke without providing a smooth transition between these points. The summary could be improved by restructuring the sentences to provide a more logical progression of information. For example, it could start by introducing the whale and the record it broke, then explain the distance it swam, and finally mention where it started its journey.
\end{tcolorbox}

\subsection{Outlier Example for Fluency} 
\begin{tcolorbox}[colback=gray!30, colframe=black]
\label{apdx_flu}
\textbf{Article Id:} \\
dm-test-4001b252a072ac149c70840b22299cc6cfab3bae\\
\textbf{Machine Summary:} radamel falcao has scored four goals all season louis van gaal 's side . united will have to pay \# 46million to make falcao 's transfer permanent . united are unlikely to take up that option . \\
\textbf{Expert Scores:} 5,5,5(avg - 5)\\
\textbf{AutoEvalScore:}  2.3\\
\textbf{Reasoning:}  \\
The sentence \textit{"radamel falcao has scored four goals all season louis van gaal 's side ."} is grammatically incorrect. It should be rephrased to something like Radamel Falcao, who is part of Louis van Gaal's side, has scored four goals all season. Similarly, in the sentence \textit{united will have to pay \# 46million to make falcao 's transfer permanent .}, the symbol '\#' is incorrect. It should be '£' to represent British Pounds."
\end{tcolorbox}

\subsection{Outlier Example for Consistency} 
\begin{tcolorbox}[colback=gray!30, colframe=black]
\label{apdx_consis}
\textbf{Article Id:} \\
d75b043ebefc3098aea84d92bb8bec0f509b1563 \\ 
\textbf{Machine Summary:} three of the militants were killed by iranian forces in the town of negur . jaish al adal claimed responsibility for the attack . the iranian state media says the militants crossed into the country . the militants have been killed in clashes with pakistan , iranian media says . the sunni muslim group says it is investigating the incident . \\
\textbf{Expert Scores:} 5,5,5 (avg - 5) \\
\textbf{AutoEvalScore:} 2.6 \\
\textbf{Reasoning:} The statement \textit{the militants have been killed in clashes with Pakistan} is not accurate according to the article. The article states that "Eight Iranian border guards have been killed in clashes with militants near the border with Pakistan" and "Three of the militants were killed by Iranian forces in the fighting Monday in the southeastern town of Negur". Therefore, the militants were not killed in clashes with Pakistan, but rather with Iranian forces. Also, the statement \textit{the sunni muslim group says it is investigating the incident} is incorrect. The group only claimed responsibility for the attack. The investigation is being conducted by the security agencies of Pakistan, not the Sunni Muslim group.
\end{tcolorbox}